\documentclass[runningheads]{llncs}
\usepackage[T1]{fontenc}
\usepackage{graphicx}
\usepackage{todonotes}
\usepackage{pifont}
\usepackage{booktabs}
\usepackage{csquotes}
\usepackage{hyperref}

\definecolor{snow}{HTML}{59B2E3}
\definecolor{snowpink}{HTML}{C5628F}
\definecolor{darksnow}{HTML}{335479}
\definecolor{snoworange}{HTML}{FF9F36}
\definecolor{snowgray}{HTML}{5E5E5E}
\definecolor{snowviolet}{HTML}{6D559E}

\usepackage{graphicx,wrapfig}

\usepackage{adjustbox}
\newcommand{\cfbox}[1]{%
    \adjustbox{cfbox=#1 1px,max width=\linewidth}%
}

\newcommand{\smolbox}{%
    \adjustbox{max width=0.22\linewidth,cfbox=darksnow 1px}%
}

\begin{document}
\title{Unchecked and Overlooked:\\Addressing the Checkbox Blind Spot in Large Language Models with CheckboxQA}
\titlerunning{Addressing the Checkbox Blind Spot in LLMs with CheckboxQA}
%
\author{Michał Turski \and Mateusz Chiliński \and Łukasz Borchmann}
\authorrunning{M. Turski et al.}
%
\institute{Snowflake AI Research\\\email{michal.turski@snowflake.com}}


%
\maketitle              
\begin{abstract}
Checkboxes are critical in real-world document processing where the presence or absence of ticks directly informs data extraction and decision-making processes. 
Yet, despite the strong performance of Large Vision and Language Models across a wide range of tasks, they struggle with interpreting checkable content. This challenge becomes particularly pressing in industries where a single overlooked checkbox may lead to costly regulatory or contractual oversights. To address this gap, we introduce the CheckboxQA dataset, a targeted resource designed to evaluate and improve model performance on checkbox-related tasks. It reveals the limitations of current models and serves as a valuable tool for advancing document comprehension systems, with significant implications for applications in sectors such as legal tech and finance.\\

The dataset is publicly available at:\\{\color{darksnow}\url{https://github.com/Snowflake-Labs/CheckboxQA}}

\keywords{Dataset \and Visual Question Answering \and Visually Rich Documents \and Document Understanding \and Information Extraction}
\end{abstract}

\begin{figure}
{\footnotesize \textbf{\color{darksnow}Question:} Does the company disclose grants exceeding \$5K? \hfill \textbf{\color{darksnow}Answer:} No}\vspace{0.5em}

\cfbox{darksnow}{\includegraphics[width=\linewidth]{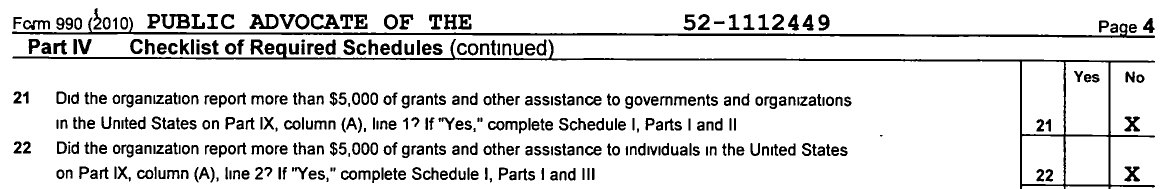}}\vspace{1em}

{\footnotesize \textbf{\color{darksnow}Question:} What vehicle type categories are recorded? \hfill \textbf{\color{darksnow}Answer:} CMV, HAZMAT}\vspace{0.5em}
\cfbox{darksnow}{\includegraphics[width=\linewidth]{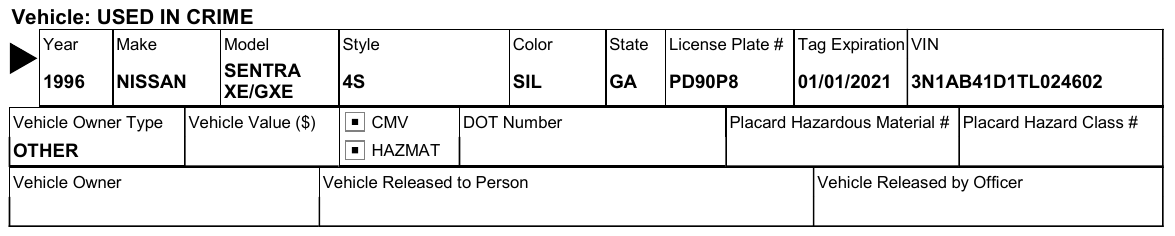}}
    \caption{CheckboxQA consists of varied questions requiring interpretation of checkable content in the context of visually rich documents. Required answers range from simple yes/no to lists of values.}
    \label{fig:examples}
\end{figure}

\section{Introduction}
Accurate checkbox interpretation is vital to organizational workflows, as any oversight can result in incorrect data entry, unaddressed legal obligations, or compliance breaches. Legal contracts, for instance, often include checkboxes to confirm acceptance of clauses like `\textit{Non-Disclosure Agreement Accepted},' while financial documents rely on them for optional selections such as `\textit{Include Life Insurance}.'  Processing errors in these small, yet significant elements can trigger substantial repercussions, ranging from data inaccuracies and legal misunderstandings to operational inefficiencies and regulatory violations.

Automation of processing documents with such elements promises substantial gains in efficiency and accuracy but requires robust visual detection capabilities and nuanced contextual understanding. Although recent advances in large vision and language models (LVLMs) have shown remarkable effectiveness in tasks spanning image classification, object detection, and text recognition, these models frequently stumble when encountering checkable content in documents. 

Several factors may contribute to this shortfall: checkboxes are typically small and visually subtle, demanding fine-grained detection; their significance often hinges on the surrounding text and overall document structure; and available training data fail to include examples that capture the intricacies of checked versus unchecked states.

\looseness=-1 In response to these shortcomings, we present CheckboxQA---a specialized dataset designed to advance Document AI capabilities. 
It comprises diverse documents annotated with question-answer pairs that hinge on accurate checkbox interpretation (Figure~\ref{fig:examples}). By focusing on this of\-ten-over\-looked facet of document processing, CheckboxQA bridges a critical gap in existing benchmarks and paves the way for more precise, robust, and context-sensitive models.

\section{Related Works}

\looseness=-1 Checkbox comprehension is a longstanding challenge in document processing. Before the deep learning era, traditional approaches relied on rule-based image analysis to locate checkbox squares through geometric heuristics and morphological operations. Once regions were detected, determining whether they were checked typically involved measuring pixel density or connectivity within those boxes. While these heuristic-driven methods were computationally efficient and required minimal labeled data, they often struggled with varied layouts and forms, necessitating extensive parameter tuning for each new layout \cite{istle2004optical,adams2006feature,6802658,zahray2019automating}.

With the advent of learning-based vision models, researchers began exploring neural networks for more robust and generalizable checkbox recognition \cite{murphy2021checkbox,nagarikar2021input,10229417,folks2024computer}. By learning features directly from data, these methods outperformed template-based systems and could handle heterogeneous checkbox styles and noisy inputs. More recently, the rapid development of LLMs has triggered a paradigm shift, unifying diverse tasks---previously addressed by specialized architectures---under a broader question-answering framework \cite{mccann2018naturallanguagedecathlonmultitask,DBLP:journals/corr/abs-1910-10683,Radford2019LanguageMA}. Document intelligence has benefited from this shift, moving towards natural language interfaces for visually rich documents \cite{10.1007/978-3-030-86331-9_47,tang2023unifyingvisiontextlayout,kim2022ocrfreedocumentunderstandingtransformer}.

In line with this trend, recent Document VQA benchmarks have emerged to promote research into visually grounded QA, including DocVQA \cite{mathew2021docvqadatasetvqadocument}, InfographicsVQA \cite{mathew2021infographicvqa}, SlideVQA \cite{tanaka2023slidevqadatasetdocumentvisual}, and DUDE \cite{vanlandeghem2023documentunderstandingdatasetevaluation}. These address a variety of question types and visual complexities; however, they do not explicitly isolate the unique challenge of interpreting checkbox fields.

The proposed CheckboxQA dataset fills this gap: It maintains the Document VQA paradigm while targeting a critical but underrepresented element---checkboxes---and thus complements existing resources by focusing on a form component they often treat implicitly or overlook entirely.

\section{CheckboxQA Benchmark}
This section introduces CheckboxQA, a curated dataset dedicated to the interpretation of checkboxes in visually rich documents. We describe how the dataset was compiled, annotated, and validated, along with key statistical insights that underscore its diversity and real-world applicability.

\begin{figure}
\centering\smolbox{\includegraphics{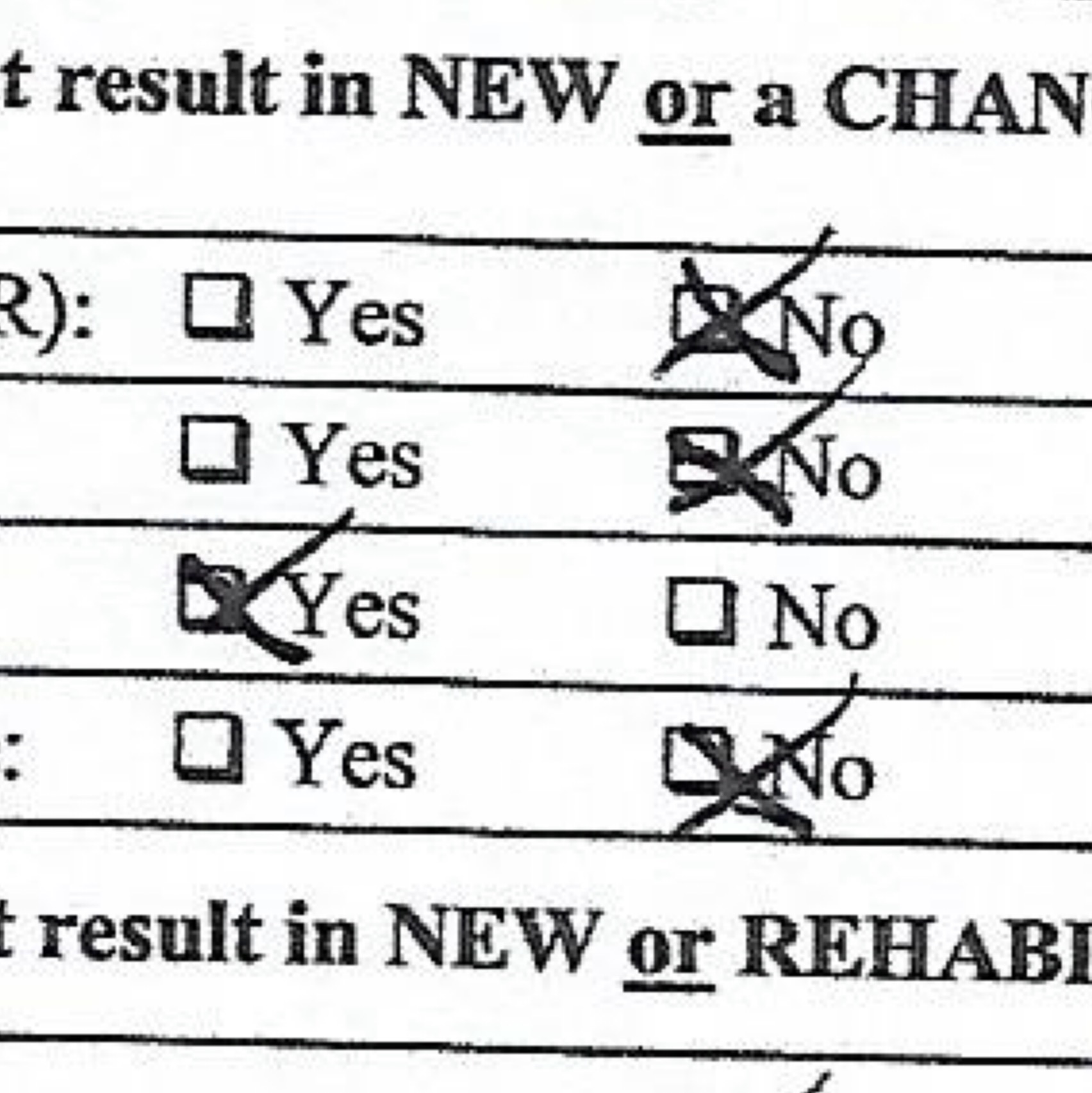}}
\smolbox{\includegraphics[width=\linewidth]{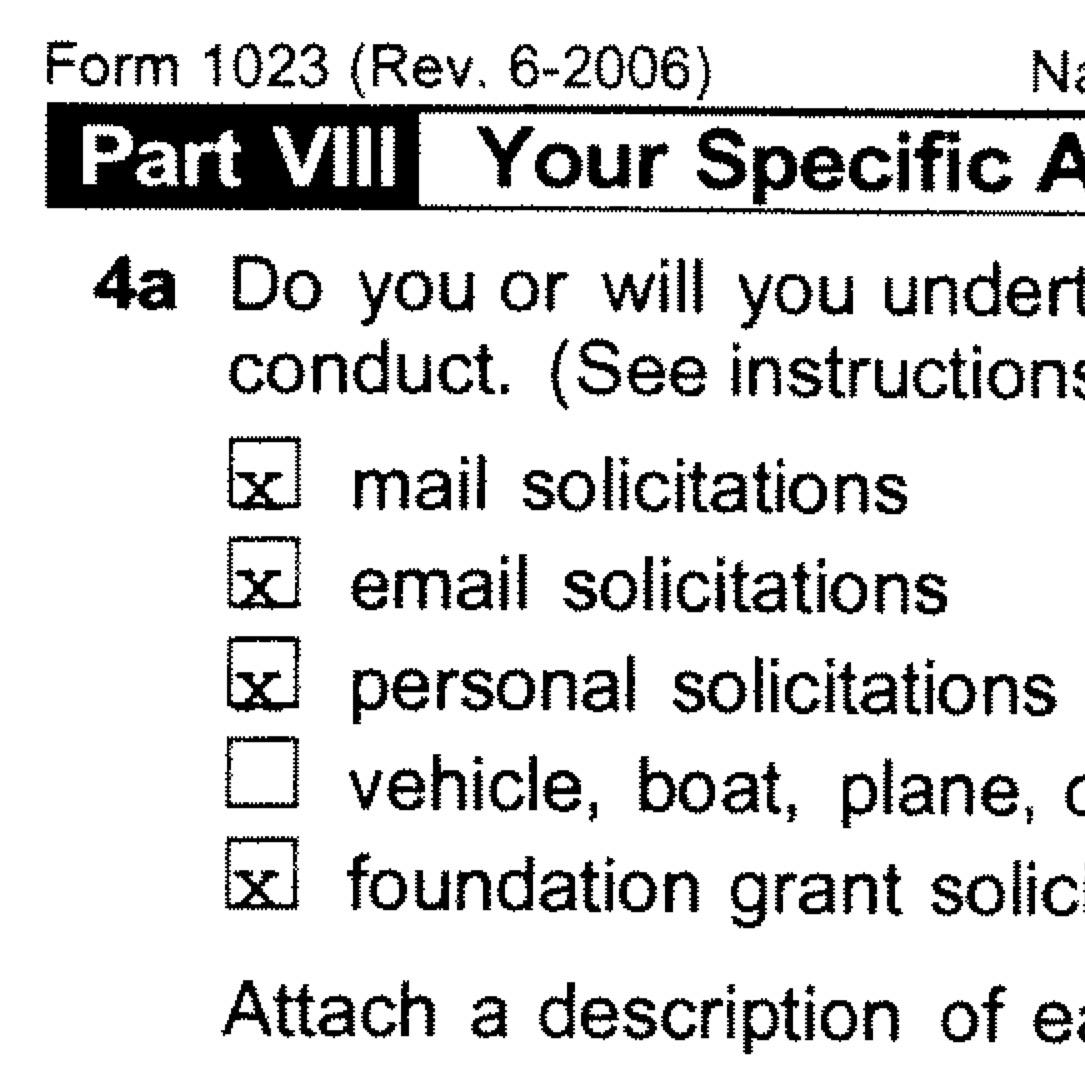}}
\smolbox{\includegraphics[width=\linewidth]{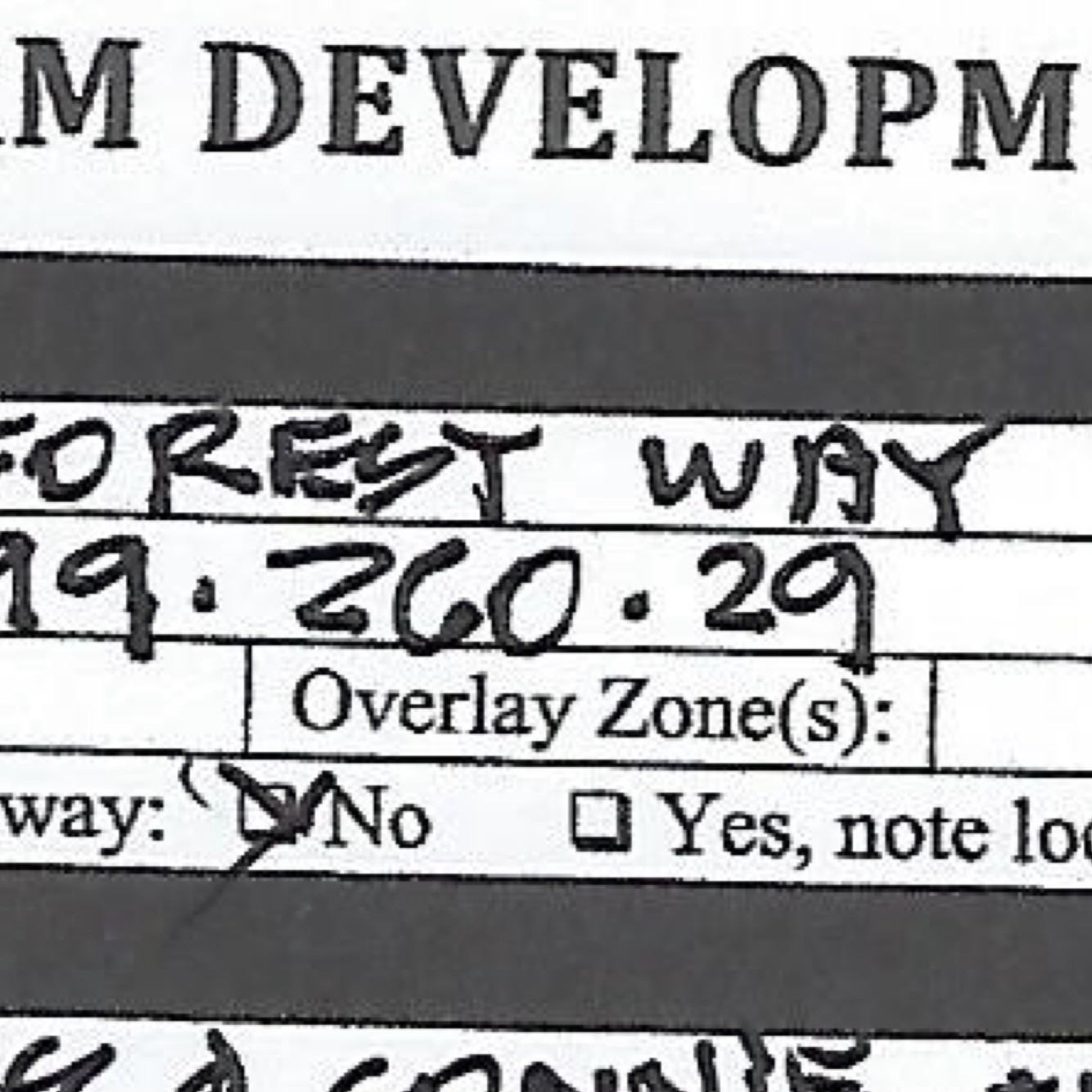}}
\vspace{0.006\linewidth}
\smolbox{\includegraphics[width=\linewidth]{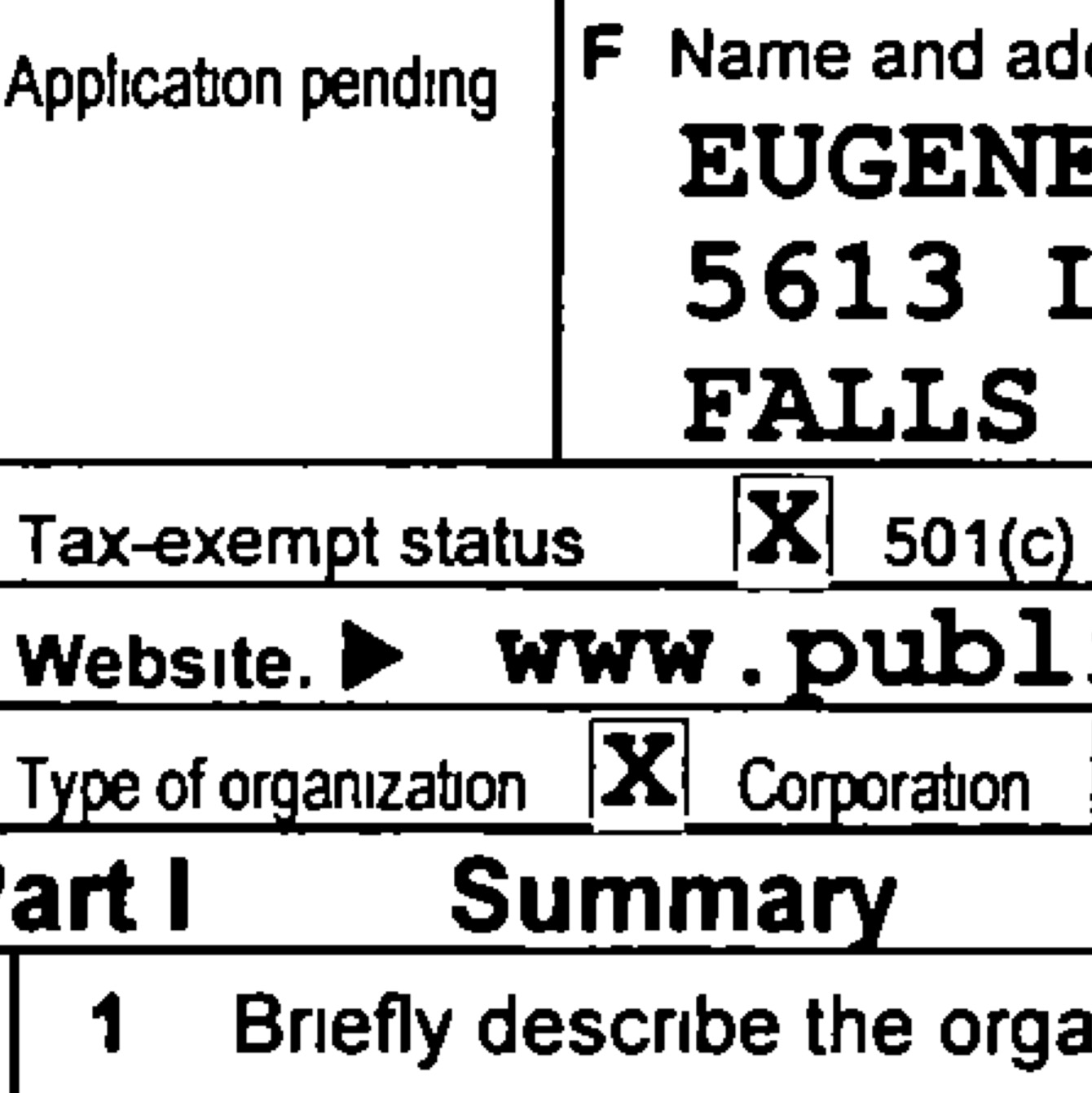}}
\smolbox{\includegraphics[width=\linewidth]{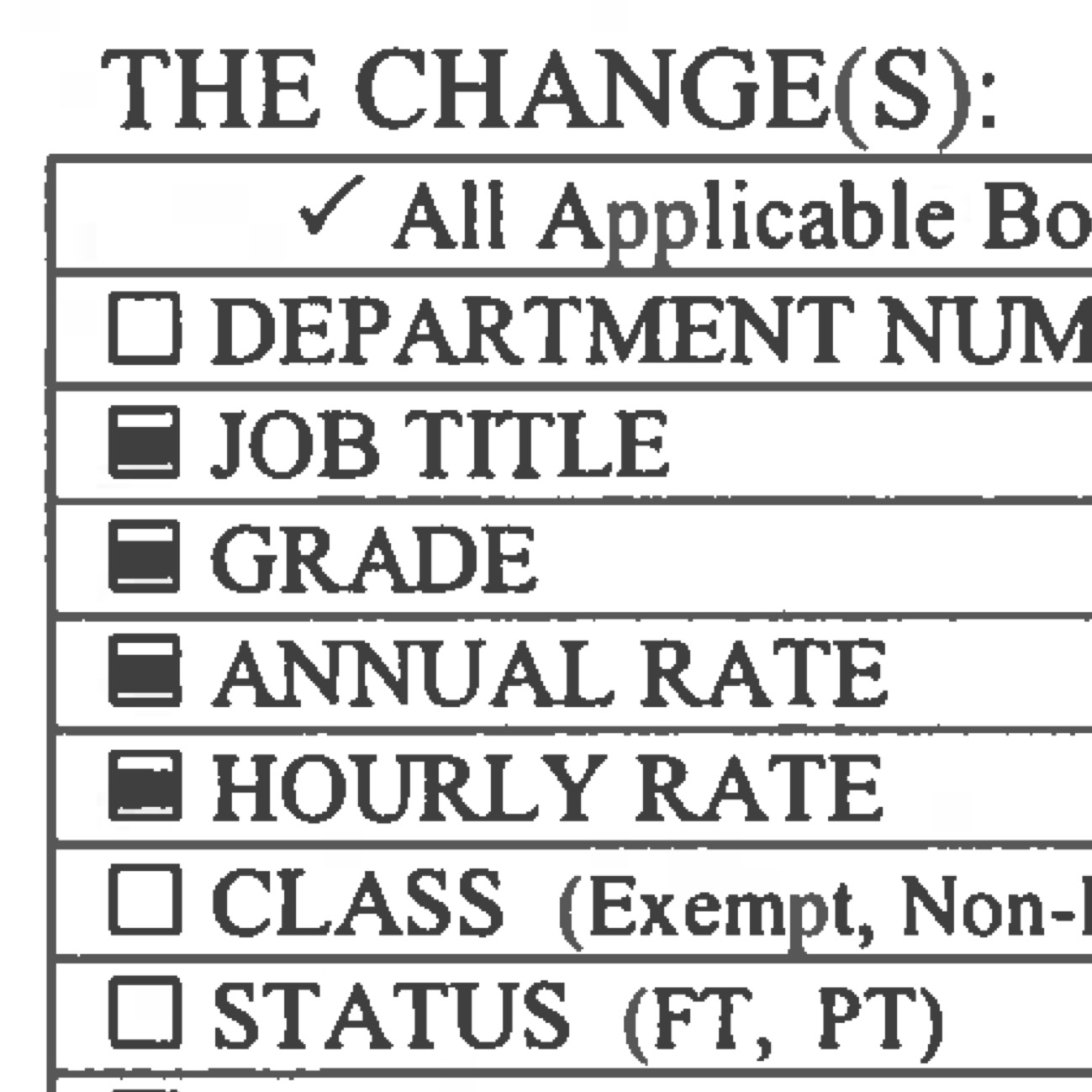}}
\smolbox{\includegraphics[width=\linewidth]{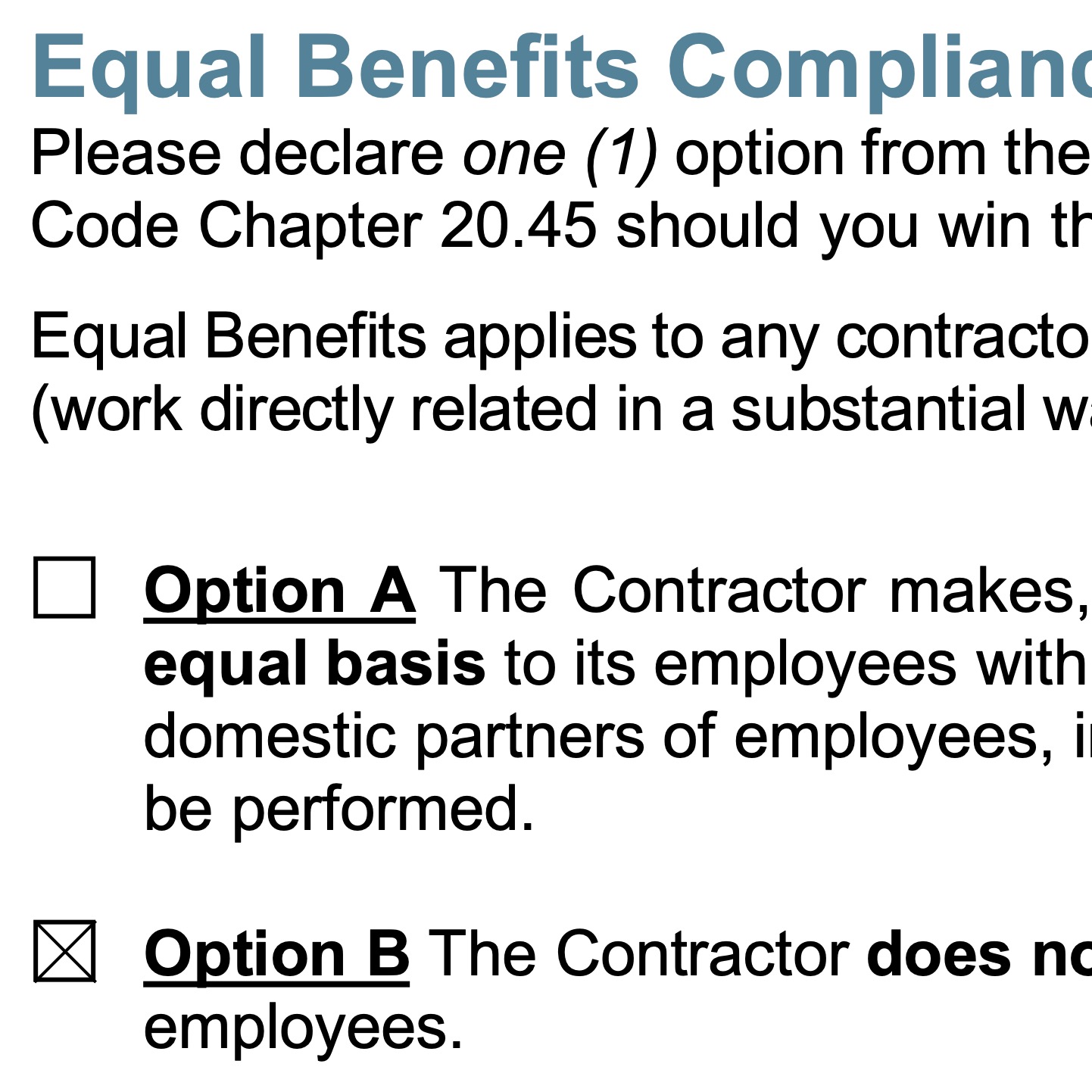}}
\smolbox{\includegraphics[width=\linewidth]{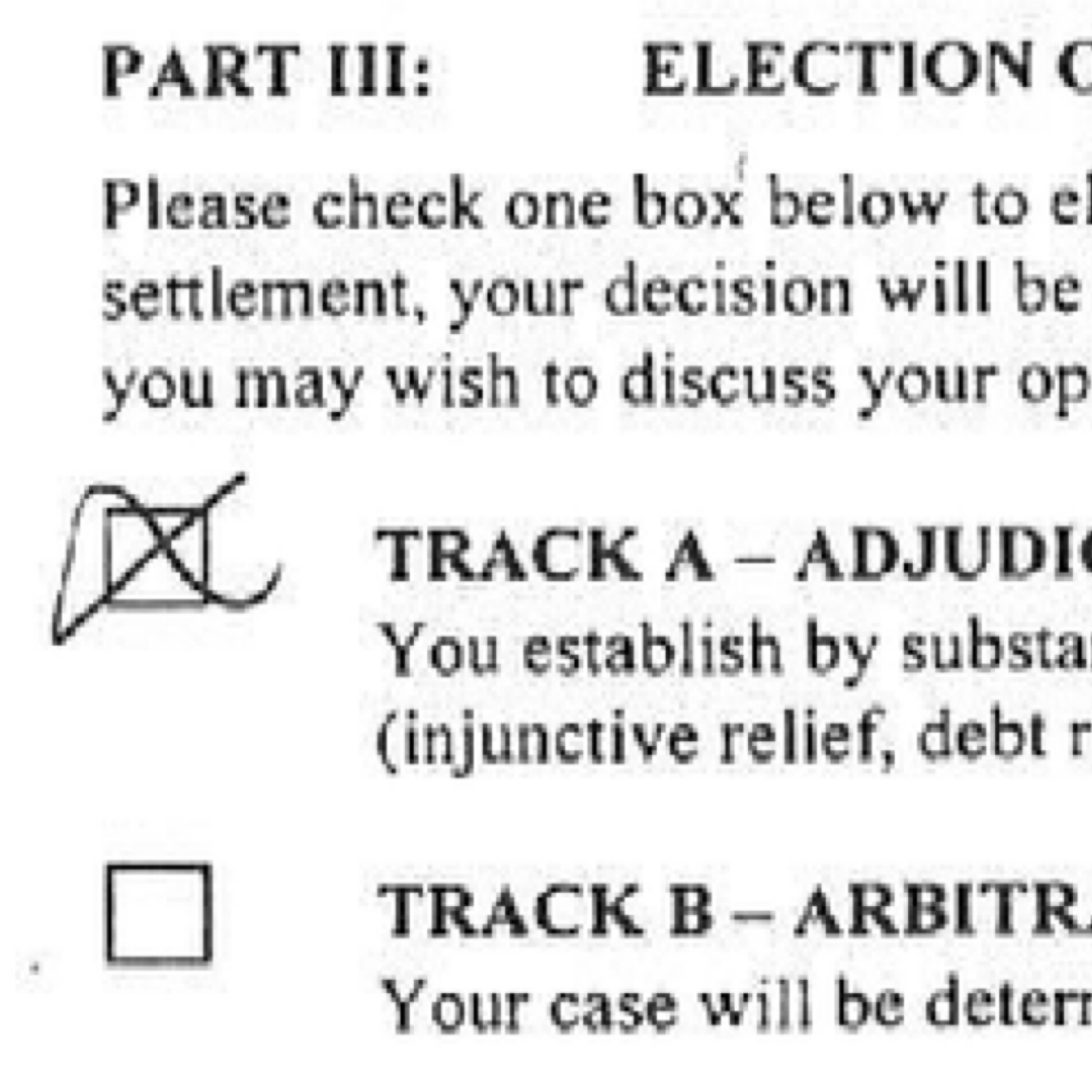}}
\smolbox{\includegraphics[width=\linewidth]{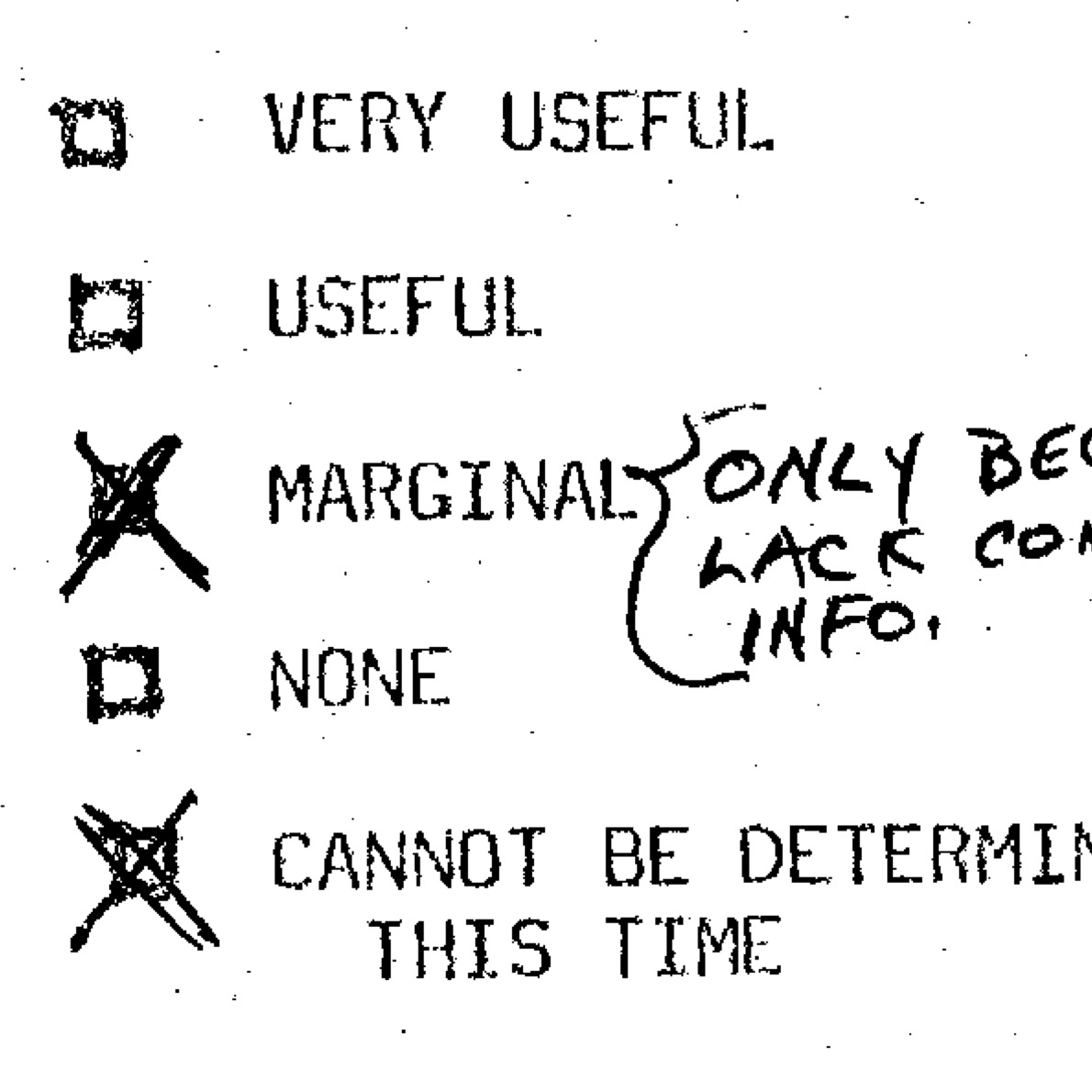}}
\caption{Excerpts from CheckboxQA documents (not an exhaustive list).}
\label{fig:document_grid}
\end{figure}

\subsection{Documents Collection}

We collected document samples from a public subset of DocumentCloud,\footnote{\url{https://www.documentcloud.org/}} ensuring a balance of form types and visual styles. Our primary selection criteria emphasized the following.

\paragraph{Presence of Checkboxes.} Each document contains one or more fields of varying shapes and sizes. Additionally, we required that at least one of the selections in the document was positive.

\paragraph{Visual Diversity.} To further ensure diversity, we cross-checked layout complexities (multi-column forms, tabular structures, single-page vs. multi-page) and document qualities (transparent vs. slightly degraded scans) to mimic real-life digitization scenarios (see Figure~\ref{fig:document_grid}). 

\paragraph{Language and License.} Only English documents published under permissive licenses 
were included.\vspace{1em}


\begin{figure}
    \centering
    \includegraphics[width=0.92\linewidth]{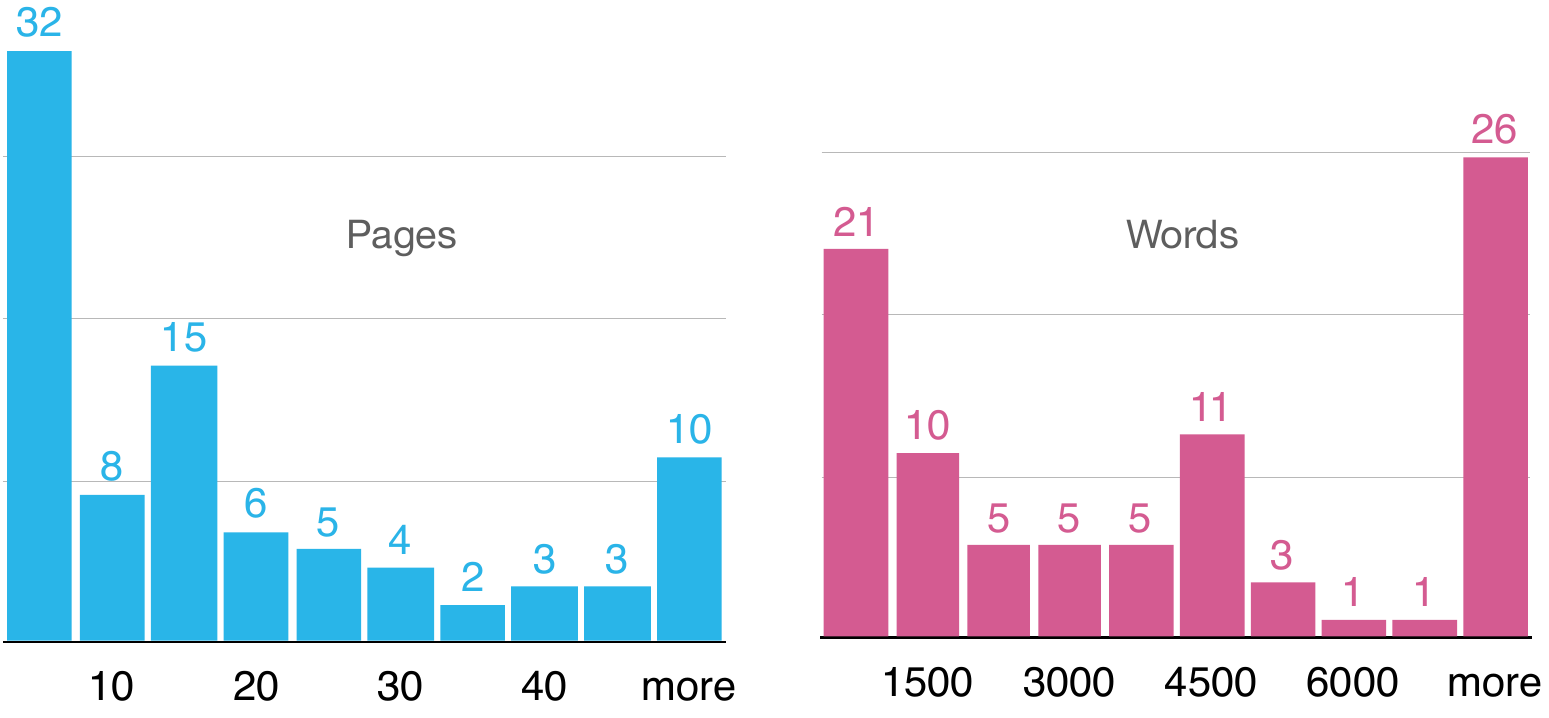}
    \caption{Histogram of collected documents lengths in terms of PDF pages and words. The plot on the right indicates a long tail of lengthy documents.}\label{fig:doc_lengths}\end{figure}

\noindent Overall, around 90 multi-page documents fulfilling these rigorous selection criteria were collected and used in the QA annotation process. Figure~\ref{fig:doc_lengths} analyzes their length in terms of the total number of pages and words.

\begin{figure}
    \centering
    \includegraphics[width=\linewidth]{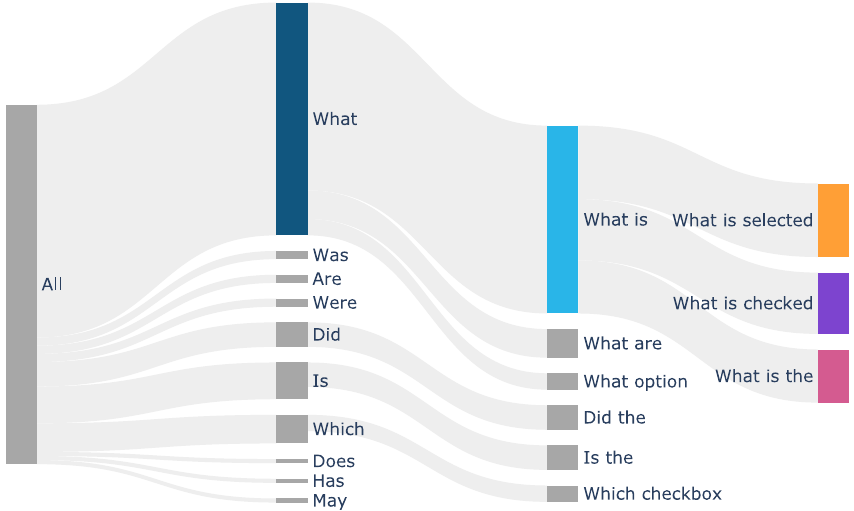}
    \caption{Most popular question prefixes.}
    \label{fig:sankey}
\end{figure}

\begin{figure}
    \centering
    \includegraphics[width=0.95\linewidth]{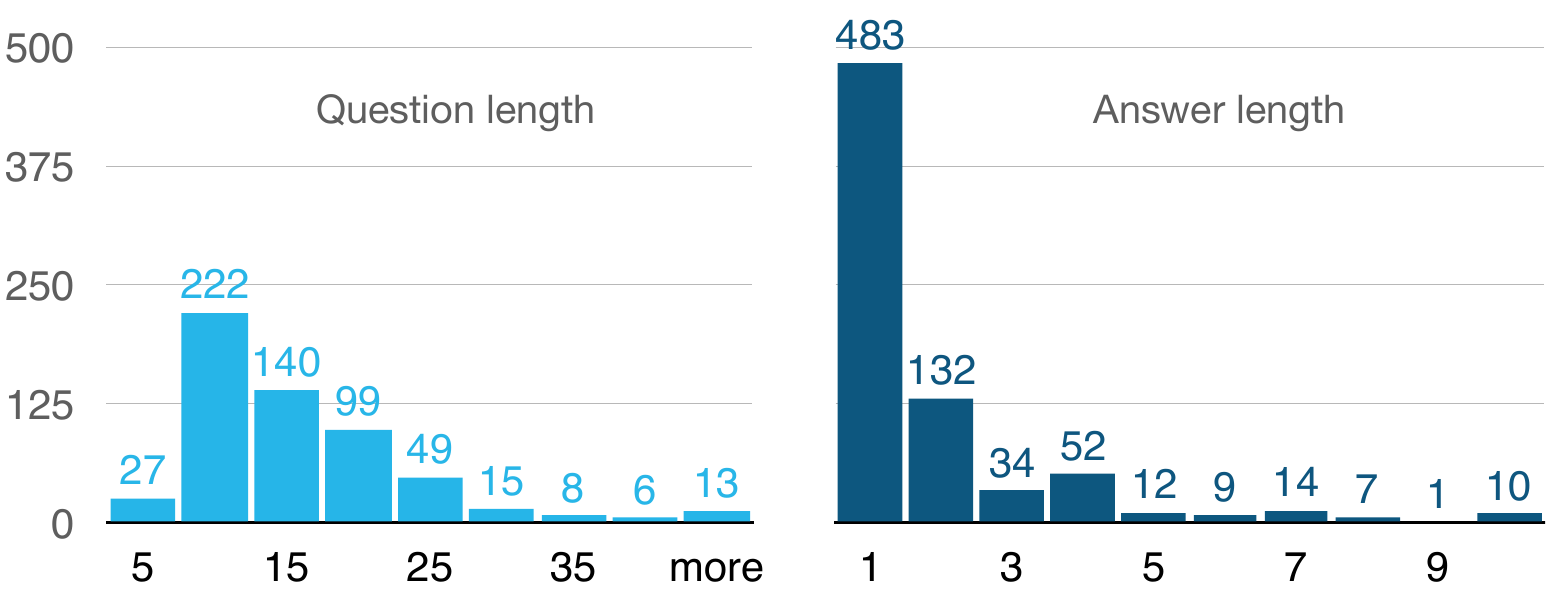}\vspace{1em}

    \caption{Histogram of annotated questions and answers lengths. 
    }
    \label{fig:questions}
\end{figure}

\subsection{Annotation of Question-Answer Pairs}
\label{sec:annotation}

CheckboxQA was constructed using guidelines adapted from a broader question-answer annotation framework by annotators experienced in creating document VQA datasets.

\setlength{\columnsep}{25pt}
\begin{wrapfigure}{r}{0.48\linewidth}
    \centering
    \includegraphics[width=\linewidth]{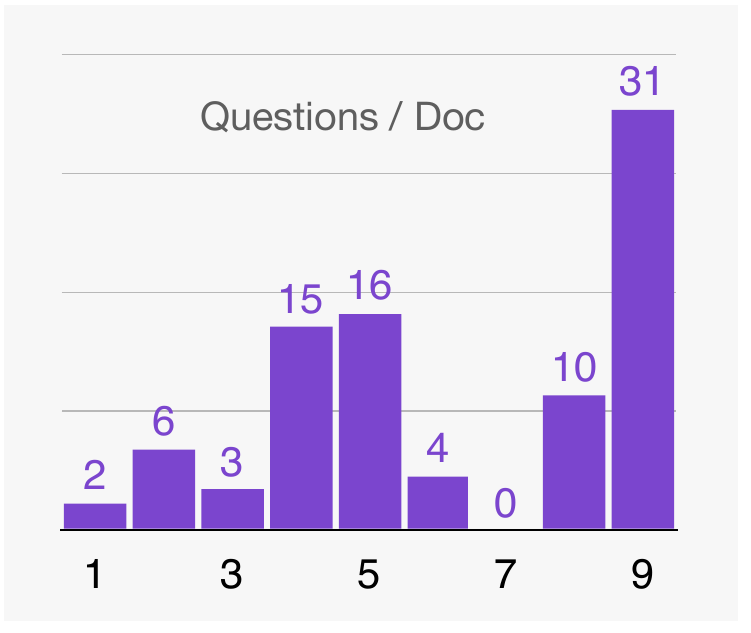}
    \caption{Distribution of QA pairs across CheckboxQA documents.}
    \label{fig:questions}
\end{wrapfigure} Over three weeks (at an estimated cost of 3K USD), contractors systematically reviewed each document and generated up to ten questions targeting checked versus unchecked items. Answers were kept concise---\textit{Yes}, \textit{No}, a single label, or a short list of labels---to pinpoint which boxes were marked without unnecessary phrasing.

The process yielded approximately 600 question-answer pairs, half requiring Yes/No answers and half being open-ended. Figure~\ref{fig:sankey} studies popular patterns of question prefixes, whereas Figure~\ref{fig:questions} presents statistics of QA lengths and their distribution.

\subsection{Problem Formulation and Experimental Setup}\label{sec:problem-formulation}

We define checkbox interpretation in a Document VQA paradigm \cite{mathew2021docvqadatasetvqadocument}. Given the document images (typically resulting from rendering a PDF file) and a question focusing on checkbox-related information, the model must produce the correct textual answer. The answer can take various forms, depending on the question:
\begin{itemize}
    \item \textbf{Binary (Yes/No):} \emph{`Is this checkbox checked?'}
    \item \textbf{Singleton Identifier:} \emph{`Which option is checked here?'} (if exactly one option is selectable)
    \item \textbf{List of Checked Items:} \emph{`Which vehicle categories are indicated as applicable?'} (if multiple options can be selected)
\end{itemize}

To succeed, a system must jointly parse textual content, identify relevant checkbox regions, and link them to the question context. This involves detecting checkbox-like elements and determining if they are checked or unchecked, reading context (surrounding text, labels, or instructions) to ground each checkbox in meaningful semantics, and answering the question accurately by fusing the checkbox state with the textual context.

\subsection{Evaluation Metrics}

Following prior work on Document VQA and Scene Text VQA, we adopt a single metric termed 
\emph{Average Normalized Levenshtein Similarity} (ANLS) to evaluate 
model predictions for CheckboxQA \cite{biten2019icdar2019competitionscene,mathew2021docvqadatasetvqadocument}.

Effectively, it is a fuzzy variant of accuracy, where string similarity above the threshold yields a partial score. A perfect string match results in \(\mathrm{ANLS} = 1\) (that is zero edit distance), while answers with a significant mismatch reduce the score accordingly. Consequently, ANLS captures minor variations in wording (small edit distance) and significant discrepancies between the predicted and ground-truth answers.

Specifically, we rely on the ANLS$^{*}$ variant of ANLS (that, among others, supports list answers in addition to plain values) with a minimal similarity threshold of 0.5 \cite{peer2024anlsuniversaldocument}.

\subsection{Baseline Approaches and Human Performance}

We evaluate a suite of baseline models to assess the difficulty of CheckboxQA and provide reference performance levels.

Evaluation of commercial LVLMs follows the previously established protocol and prompts \cite{borchmann2024notesapplicabilitygpt4document}. Specifically, we convert PDFs into a series of PNG images having 2048px along longer dimension,\footnote{In rare cases where given this size, the model couldn't fit the entire context, the longer dimension was reduced to 1024px or 768px.} and feed them to the model with question preceded by short instruction:

\begin{displayquote}
\footnotesize
\texttt{Answer the question. Do not write a full sentence. Provide a value\\as a Python list. If there is a single answer, the output should\\
be a one-element list like ["ANSWER"]. If there are multiple valid\\answers, the list will have several elements, e.g., ["ANSWER 1",\\"ANSWER 2"]. Question:}
\end{displayquote}
Open-source LVLMs are evaluated using vLLM \cite{kwon2023efficientmemorymanagementlarge} 
with default inference options.

Finally, we employ human annotators to gauge the upper bound of CheckboxQA performance. It was obtained by passing all of the documents and assigning questions to a different annotator in precisely the same way models see them during the inference.

\section{Results and Analysis}

We conducted experiments with state-of-the-art commercial and open-source solutions, including models from GPT-4o \cite{openai2024gpt4ocard}, Gemini 2.0 \cite{geminiteam2024geminifamilyhighlycapable}, Qwen 2.5 VL \cite{wang2024qwen2vlenhancingvisionlanguagemodels}, and Pixtral \cite{agrawal2024pixtral12b} families to evaluate how well large vision-language models handle the fine-grained task of interpreting checkboxes.

\subsection{Quantitative Analysis}\label{sec:quantitative-analysis}

Table~\ref{tab:results} reports the performance of various LVLMs and a human baseline. Among the tested systems, Qwen 2.5 VL 72B attains the highest score at 83.2\%, significantly outperforming GPT-4o. Smaller Qwen variants, Pixtral 12B, and the Gemini series exhibit more modest results in the range of 43.6\% to 71.9\%. GPT-4o mini remains at the lower end with a score of 40.4\%.

Notably, the top Qwen models perform relatively well, which may suggest that their pretraining data includes a substantial number of form-like images with checkbox annotations. If so, it substantiates our core claim: checkbox content has generally been overlooked in conventional large-scale training, and models that happen (by design or otherwise) to include such specialized examples gain a distinct advantage in tasks like those included in CheckboxQA.

Despite these advances, every model still falls short of the near-ceiling human baseline of 97.5\%, underscoring the difficulty of accurately identifying and interpreting checkmarks. These elements are often visually subtle or positioned unpredictably, demanding fine-grained spatial and textual reasoning that remains challenging for LVLMs.

\begin{table}[t]
    \caption{CheckboxQA evaluation results, compared to human performance.}

    \centering
    \setlength{\tabcolsep}{12pt}
    \def\arraystretch{1.1}
    \begin{tabular}{lc}
    \toprule
    \textbf{Model} & \textbf{Score} (ANLS$^{*}$) \\
    \midrule
     Qwen 2.5 VL 72B & 83.2 \\
     Qwen 2.5 VL 7B & 71.9 \\
     Snowflake Arctic-TILT 0.8B {\color{snowgray} \scriptsize 2025-03 } & 66.8 \\
     GPT-4o {\color{snowgray} \scriptsize 2024-11-20}    & 66.7 \\
     Gemini 2.0 Pro {\color{snowgray} \scriptsize exp-02-05} & 59.7 \\
     Gemini 2.0 Flash Lite {\color{snowgray} \scriptsize preview-02-05} & 55.2 \\
     Pixtral 12B & 56.9 \\
     Gemini 2.0 Flash {\color{snowgray} \scriptsize 001} & 54.4 \\
     Qwen 2.5 VL 3B & 43.6 \\
     GPT-4o mini {\color{snowgray} \scriptsize 2024-07-18} & 40.4 \\
    \midrule
    human performance & 97.5 \\
    \bottomrule
    \end{tabular}
    \label{tab:results}
\end{table}

\subsection{Qualitative Observations}

Qualitative assessment of CheckboxQA reveals specific scenarios where these models systematically fail. Below, we highlight examples from leading models to illustrate recurring mistakes.

\begin{figure}
    {\footnotesize \textbf{\color{darksnow}Question:} Is additional tasking required?\\
    \textbf{\color{darksnow}GPT-4o, Gemini:} Yes}\vspace{0.5em}\\

    \centering\cfbox{darksnow}{\includegraphics[width=0.65\linewidth]{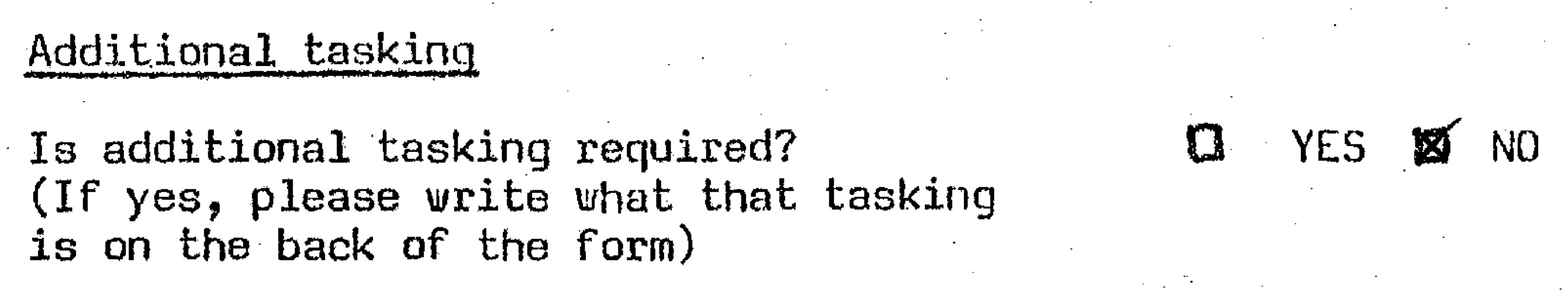}}\vspace{2em}

    \raggedright{\footnotesize \textbf{\color{snow}Question:} What option is selected in `Education: minimum level required'? \\ \textbf{\color{snow}GPT-4o:} Bachelor's}\vspace{0.5em}\\

    \centering
    \cfbox{snow}{\includegraphics[width=\linewidth]{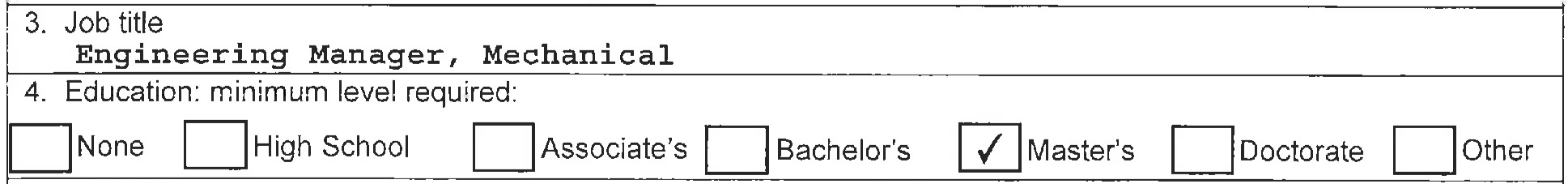}}
    \caption{While visually, the label for the checkbox is clearly anchored to a different answer, strong LVLMs incorrectly associate the text with the checkbox.}    \label{fig:missaligned}
\end{figure}

\paragraph{Misaligned Checkbox and Text Context.} In certain documents, the checkbox for a given label can appear on either the left or right side of the text. Some LVLMs fail to associate the correct checkbox with its label (Figure~\ref{fig:missaligned}).

\paragraph{Defaulting to Textual Clues Instead of Checkbox States.}
When asked a binary question about additional tasks, models sometimes rely on textual context rather than checking the actual box state.

\begin{figure}[t]
    {\footnotesize \textbf{\color{darksnow}Question:} What are the types of products to benefit from use of reported information? \\ \textbf{\color{darksnow}GPT-4o, Gemini:} BASIC INTELLIGENCE, CURRENT INTELLIGENCE, ESTIMATIVE INTELLIGENCE, S\&T INTELLIGENCE}\vspace{0.5em}\\

    \centering
    \cfbox{darksnow}{\includegraphics[width=0.45\linewidth]{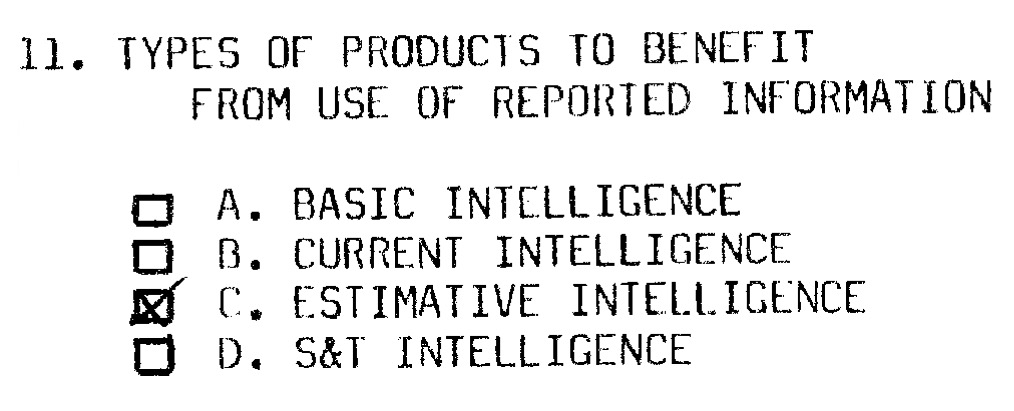}}

    \caption{Models commonly fail to discriminate among checked and unchecked boxes and instead enumerate all available labels.}  \label{fig:list}
\end{figure}

\paragraph{Selecting All Possible Options in a List.}
In scenarios where only one or a few checkboxes should be selected, models occasionally list every available option (Figure~\ref{fig:list}).

\paragraph{Ignoring Table Structure for Checkbox Fields.}
When checkboxes are placed in a table, the surrounding text may become distracting for the model.

\paragraph{Returning Special Symbols Instead of Textual Answers.}
Certain models respond with literal symbols (e.g., \texttt{X} or \ding{51}) in place of text.

\paragraph{Confusing the Question Text as the Answer.}
Another frequent error occurs when the model interprets the question text as part of the response.\vspace{1em}

Overall, these failures underscore the need for robust representations that jointly model the spatial arrangement of checkbox fields and the visual distinction between checked and unchecked states. They also demonstrate that large language models---even ones with robust text understanding---still struggle when the question hinges on a subtle visual or structural cue rather than text alone.

\section{Limitations}
Although CheckboxQA advances the study of checkbox-related tasks in Document AI, it remains subject to several limitations.
First, all documents and annotations are in English, potentially restricting the dataset's applicability to other languages and character sets.
Second, despite efforts to diversify the document collection, certain domains (e.g. medical or highly technical forms) may be underrepresented, limiting the dataset's coverage of specialized use cases.
Finally, while the results provide insights into model performance, they focus on a specific subset of commercial and open-source models. 


\section{Summary}

We presented CheckboxQA, a targeted dataset designed to evaluate how large vision-language models handle checkboxes in visually rich documents.
This task is of considerable practical importance, given that checkbox errors can lead to significant operational missteps. For instance, a missed opt-out box in a legal contract could expose a firm to privacy breaches, underscoring how even minor checkable fields can carry significant practical consequences.

While large vision-language models have made substantial strides in document understanding, accurately interpreting checkboxes remains a significant challenge. Even the top-performing models in our experiments fell substantially short of human-level performance, indicating persistent gaps in fine-grained visual reasoning and layout comprehension. These gaps are particularly evident in misaligning checkboxes with the appropriate text, defaulting to textual cues when visual inspection is required, and failing to filter out unchecked items in multi-selection scenarios.

Ultimately, the performance trends observed in CheckboxQA suggest that progress in broad document understanding does not uniformly translate to proficiency in micro-level visual tasks. By isolating the challenges posed by checkboxes, our benchmark aims to catalyze research on more specialized form understanding methods, paving the way for systems that can handle all the intricate details of real-world forms with high accuracy.

\clearpage
\appendix
\footnotesize
\section{Dataset Card for CheckboxQA}

\paragraph{Dataset Overview.}
CheckboxQA is a specialized benchmark focused on interpreting checkboxes in visually rich documents. It comprises multi-page English documents containing one or more checkboxes and about 600 question-answer (QA) pairs. The dataset provides an in-depth look at model capabilities for checkbox detection and state interpretation within real-world forms.

\paragraph{Motivation.}
Although Document AI benchmarks are abundant, most do not isolate the challenge of distinguishing checked vs. unchecked states. CheckboxQA addresses this gap, ensuring models accurately associate checkbox states with nearby textual descriptions.

\paragraph{Data Collection.}
Underlying PDF files were gathered from \url{http://documentcloud.org}, emphasizing diverse form layouts (multi-column, tabular, single/multi-page) and variations in scan quality (clear vs.\ mildly degraded).

\paragraph{Language.}
All documents are in English, reflecting the prevalent use of English-language forms and permits in public-domain sources. 

\paragraph{Annotation Process.}
Trained annotators generated up to ten question-answer pairs per document, each focusing on checkbox state or label interpretation. Answers typically take one of the following forms:
\begin{itemize}
    \item Yes/No (binary),
    \item Single selection,
    \item List of selections.
\end{itemize}

\paragraph{Dataset Composition.}
\begin{itemize}
    \item \textit{Document count}: 88.
    \item \textit{Total QA pairs}: 579.
\end{itemize}

\paragraph{Intended Use.}
CheckboxQA is designed for:
\begin{itemize}
    \item Benchmarking vision-language models on fine-grained checkbox detection,
    \item Research on layout-aware document understanding,
    \item Testing end-to-end systems combining OCR, layout parsing, and QA.
\end{itemize}

\paragraph{Licensing and Distribution.}
All documents were sourced under permissive or public-domain licenses. We do not rehost files but provide a script to download them from the original providers instead. We release annotations on Apache 2.0.

\paragraph{Limitations.}
\begin{itemize}
    \item Focused on English-language documents only.
    \item Primarily covers forms, surveys, and agreements, which may not generalize to all document domains.
    \item Annotator biases or small sample sizes could limit coverage of rare checkbox designs.
\end{itemize}

\normalsize

\bibliographystyle{splncs04}
\bibliography{bibliography}
\end{document}